\documentclass[utf8]{arxiv}

\usepackage{url,lineno,microtype}
\usepackage[onehalfspacing]{setspace}
\usepackage{standalone}
\usepackage{tikz}
\usetikzlibrary{bayesnet}
\usetikzlibrary{arrows.meta} 
\usepackage{amssymb}
\usepackage{multirow}
\usepackage{booktabs} 
\usepackage{changes}

\usetikzlibrary{arrows,positioning,automata}

\usepackage[colorlinks=true, allcolors=blue]{hyperref}

\usepackage[autostyle=true, english=american]{csquotes}
\MakeOuterQuote{"}

\renewcommand{\thefootnote}{\fnsymbol{footnote}}
\renewcommand*{\thefootnote}{\dagger}

\def\keyFont{\fontsize{8}{11}\helveticabold }
\def\firstAuthorLast{Y. Li {et~al.}} %
\def\Authors{Yin Li\,$^{1}$\thanks{These authors contributed equally to this work.}, Yu Xiong\,$^{2*}$, Wenxin Fan\,$^{3}$, Kai Wang\,$^{1}$, Qingqing Yu\,$^{1}$, Liping Si\,$^{4}$, Patrick van der Smagt\,$^{5,6}$, Jun Tang\,$^{1\star}$ and Nutan Chen\,$^{6}$}
\def\Address{$^{1}$ Department of Otorhinolaryngology, The First People's Hospital of Foshan, China \\
$^{2}$ Department of Otorhinolaryngology, The Second Affiliated Hospital of Guizhou University of Traditional Chinese Medicine, Guiyang, China\\
$^{3}$ Chinese Academy of Sciences Shenzhen Institutes of Advanced Technology, Shenzhen, China \\
$^{4}$ Department of Radiology, Zhongshan Hospital, Fudan University, Shanghai, China \\
$^{5}$ ELTE University, Budapest, Hungary\\
$^{6}$ Machine Learning Research Lab, Volkswagen Group, Munich, Germany}
\def\corrAuthor{Jun Tang}
\def\corrEmail{fsyyytj@126.com}

\begin{document}
\onecolumn
\firstpage{1}

\title[Sequential Model for Predicting Patient Adherence]{Sequential Model for Predicting Patient Adherence in Subcutaneous Immunotherapy for Allergic Rhinitis} 

\author[\firstAuthorLast ]{\Authors} %
\address{} %
\correspondance{} %

\extraAuth{}%

\maketitle

\begin{abstract}
Objective: Subcutaneous Immunotherapy (SCIT) is the long-lasting causal treatment of allergic rhinitis (AR). How to enhance the adherence of patients to maximize the benefit of allergen immunotherapy (AIT) plays a crucial role in the management of AIT. This study aims to leverage novel machine learning models to precisely predict the risk of non-adherence of AR patients and related local symptom scores in three years SCIT.

Methods: 
The research develops and analyzes two models, sequential latent-variable model (SLVM) of Stochastic Latent Actor-Critic (SLAC) and Long Short-Term Memory (LSTM)
evaluating them based on scoring and adherence prediction capabilities.

Results: Excluding the biased samples at the first time step, the predictive adherence accuracy of the SLAC models is from 60\% to 72\%, and for LSTM models, it is 66\% to 84\%, varying according to the time steps. The range of Root Mean Square Error (RMSE) for SLAC models is between 0.93 and 2.22, while for LSTM models it is between 1.09 and 1.77. Notably, these RMSEs are significantly lower than the random prediction error of 4.55.

Conclusion: We creatively apply sequential models in the long-term management of SCIT with promising accuracy in the prediction of SCIT nonadherence in AR patients. While LSTM outperforms SLAC in adherence prediction, SLAC excels in score prediction for patients undergoing SCIT for AR. The state-action-based SLAC adds flexibility, presenting a novel and effective approach for managing long-term AIT.

\renewcommand{\thefootnote}{}
\footnotetext{\textit{Frontiers in Pharmacology}, research topic: Methods and Metrics to Measure Medication Adherence.}
\renewcommand{\thefootnote}{\arabic{footnote}}

\tiny
 \keyFont{\section{Keywords: Allergic rhinitis, Allergen immunotherapy, Adherence, Sequential model, Latent variable model}} %

\end{abstract}

\section{Introduction}

Allergic rhinitis (AR) is characterized by allergen-specific IgE-mediated inflammation in upper respiratory inflammation with a prevalence of up to 30\% worldwide \citep{meltzer2016allergic}. In addition to allergen avoidance as the superior criterion, allergen-specific immunotherapy (AIT) aims to induce specific allergen immune tolerance, consequently achieving a status of clinical symptom remission. The repeatable intake of the specific unmodified or chemically modified allergens (allergoids)  was the key to maintaining the symptoms. Among these approaches of AIT, subcutaneous immunotherapy (SCIT), sublingual immunotherapy (SLIT), and lymphatic immunotherapy (LIT) are demonstrated as the mainstream treatments regarding efficacy, safety, and side effects. 
Compared to the SLIT, SCIT is a clinic-dependent treatment in which the patient accepted an allergen extract injection subcutaneously. 
It is divided into the initial treatment stage (dose accumulation stage) and the maintenance treatment stage (dose maintenance stage). The World Allergy Organization (WAO) recommends that immunotherapy be maintained for three to five years and clinically recommended for at least two years. Patient adherence is a critical factor in ensuring long-lasting efficacy and sustaining symptom-relieving effects.

Due to the long duration of SCIT, cumbersome process, slow onset, individual differences in treatment effect, and other factors fundamentally impact the completeness of therapeutics. From the reported studies on AIT, the rate of adherence ranged from around 25\% to over 90\% \citep{passalacqua2013adherence}. The World health Organization (WHO) adopted the definition of "adherence" as “the extent to which a person’s behavior, such as taking medication, following a diet, or executing lifestyle changes, corresponds with agreed recommendations from a health care provider” \citep{world2003adherence}. In recent European Academy of Allergy and Clinical Immunology (EAACI) guidelines, it is highlighted to educate patients on how immunotherapy works and on explaining the importance of compliance to the regular doses for three years of treatment \citep{roberts2018eaaci}.

The multiple approaches were introduced into the field of improving adherence and supervising patient outcomes with systematic and technological interventions to prevent incomplete discontinuation of the treatment.
The intervention from a clinic in advance running through the whole treatment cycle was approved as an effective approach. 
In facing the multitude of personalized data from patients, how to precisely identify and assess the risk of upcoming non-adherent behavior, a clinical prediction model is promising in the application.

In healthcare, machine learning, especially sequential models, stands at the forefront of innovation, providing new ways to analyze complex medical data and improve patient treatments. 
Previous research primarily concentrated on non-sequential prediction methods for adherence \citep{mousavi2022determining, wang2020applying,warren2022using,ruff2019role}. This approach presents a significant limitation in treatment processes, particularly for immunotherapy that often spans extended periods, such as three years. These non-sequential methods tend to predict only the overall outcome, overlooking the intricacies of intermediate time steps. To facilitate earlier intervention, a sequential model capable of making predictions at any given time step would be markedly more beneficial. While some subsequent studies have introduced sequential models \citep{hsu2022medication,singh2022deep,schleicher2023prediction}, their scope was restricted to predicting adherence alone. Our study enhances this approach by incorporating a state-action model, which can predict both adherence and score/state. This advancement allows for more precise and detailed management of AR patients by allergologists.
These models excel in processing and analyzing time-dependent data, making them ideal for predicting patient adherence to treatments like SCIT for AIT. By effectively using sequential data, these algorithms uncover temporal patterns and correlations, leading to more accurate and personalized treatment plans.

In this study, in order to introduce the appropriate prediction model into long-period immunotherapy to customize the management of interventions and incorporate patient feedback, we have selected and evaluated two specific sequential models tailored to this scenario. Our findings demonstrate that these models are not only effective in predicting patient adherence to medical treatments but also invaluable in enhancing treatment strategies, thereby making a significant contribution to patient-centered healthcare.

\section{Methods}

\subsection{Study design}
\begin{figure*}[ht!]
    \centering
    \includegraphics[width=1.0\textwidth]{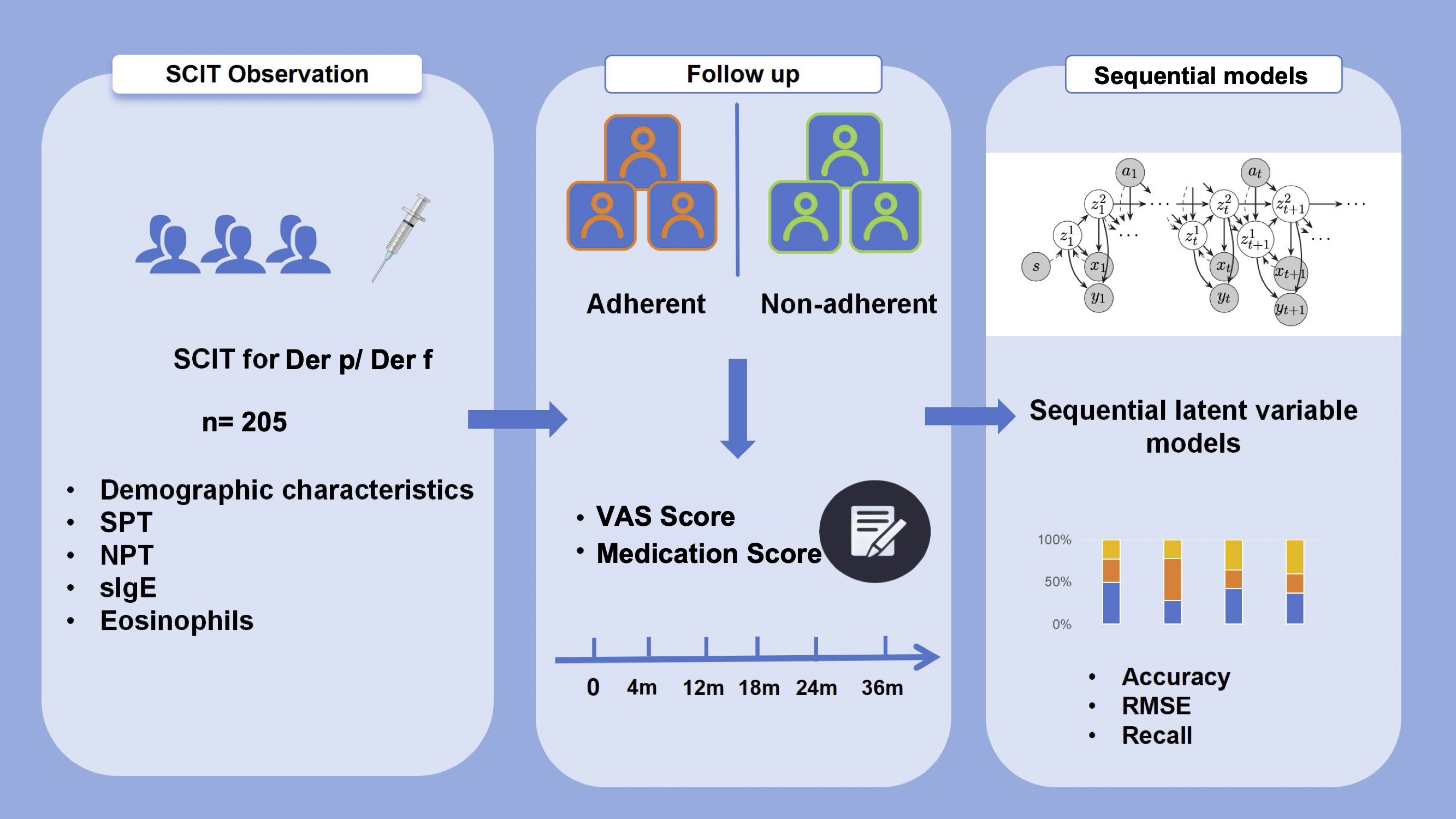}
    \caption{
Flowchart on 205 patients treated with SCIT for Der p/Der f allergy during a 36-month treatment period. The adherence was assessed with sequential latent variable models focusing on patient's demographic characteristics and clinical follow-up data.
}
    \label{fig:graphic}
\end{figure*}

The study design is a critical component that shapes the direction and reliability of our research. It includes a systematic approach to selecting the study population, the treatment methods applied, and the evaluation criteria (see Figure~\ref{fig:graphic}).

\subsubsection{Population}
A retrospective analysis including 205 AR patients who started SCIT treatment between August 2018 and September 2019 in the Immunotherapy Center at the First People's Hospital of Foshan was performed. According to the Guidelines for the Diagnosis and Treatment of Allergic Rhinitis (2015 Edition), the recruit criteria were formulated: Patients with skin index (SI) of skin prick test (SPT) ++ or above, or specific Immunoglobulin E(sIgE) level in serum to 
dermatophagoides pteronyssinus (Der p) and/or dermatophagoides farinae(Der f) i.\,e., 
$\mathrm{Der\,p} / \mathrm{Der\,f} \geq 0.35\ \text{kU/L}$,
which exposure to dust mites was confirmed as the major allergen by allergen tests, including:
(1) Patients with mild to moderate asthma;
(2) Patients with moderate to severe persistent rhinitis;
(3) Mild to moderate asthma with allergic rhinitis (and/or allergic conjunctivitis);
(4) Patients with mild to moderate asthma and eczema.
Exclusion criteria included:
(1) severe or uncontrolled bronchial asthma with continuous monitoring of Forced Expiratory Volume in one second (FEV1) \(< 70\%\) per of the expected value;
(2) Patients with asthma whose symptoms or reduced lung function continue to fail to be controlled with grade 4 or 5 treatment;
(3) Patients sensitized to other allergens such as pet furs, pollens or molds;
(4) Patients who are taking beta-2 blockers or angiotensin-converting enzyme inhibitors;
(5) Patients with serious underlying diseases, including cardiovascular and cerebrovascular diseases, autoimmune diseases and immunodeficiency diseases, malignant diseases, and chronic infectious diseases;
(6) Patients with serious mental illness, lack of compliance, or inability to understand the risks and limitations of treatment. The patients data were anonymized before use.

The study protocol was approved by the Ethics Committee of the First People's Hospital of Foshan, Foshan, China. All methods were performed 
in accordance with the relevant guidelines and regulations.

\subsubsection{SCIT treatment and evaluation}
\label{sec:method_treatment}
Before administering SCIT to enrolled patients, they will first perform a routine physical examination, inquire about related information since the last injection (including allergy symptoms), and post-injection, patients are observed for 30 minutes in case of the occurrence of side effects. Standardized adsorbed Der p and Der f allergen extracts (Allergopharma, Reinbeck, Germany) were used for SCIT. According to the manufacturer's instructions, in the dose accumulation phase with weekly injections of allergen extracts with a gradually increased concentration from 100\,SQ-U/mL to 10,000\,SQ-U/mL, respectively injected 0.2, 0.4, 0.8\,mL; after reaching the maintenance dose, 100,000 standardized quality units was used. In the maintenance phase, an injection interval of 6$\pm$2 weeks was carried out according to the manufacturer's recommendations. 

Patients receive regular treatment evaluations, including symptom scores and medication scores. The symptom score recorded a total of nasal symptoms (nasal itching, sneezing, rhinorrhea, nasal congestion), ocular symptoms (ocular itching, lacrimation), and pulmonary symptoms (shortness of breath, tightness in chest, perennial cough, wheezing), and assessed symptom severity using the visual analogue scale (VAS). In the VAS symptom score, the score of each symptom is from 0 to 10. 0 indicates that the patient has no discomfort and 10 indicates that the patient is extremely uncomfortable. The patient gives the score of each symptom according to the actual situation, and the sum of all symptom scores is the symptom score. Medication score recorded the use of current adjuvant medication within 1 month to reach symptom relief. The use of oral antihistamines, antileukotrienes, and bronchodilators were recorded as one point, local glucocorticoids as two points, oral glucocorticoids or combined medication (hormones and 
$\beta 2$
receptor agonists) as three points, and the total cumulative score was the medication score. Symptom scores and medication scores were assessed once at registration of SCIT and then thereafter.

  Due to the separated injection regimen within 16 weeks and thereafter, all the chosen patients completed the four months of SCIT, we chose the fourth month as the starting point of the observation. According to our previous experience, one year after the start was the peak of the withdrawal, so we added a time point at 18 months to further assess and follow up on the related symptom score and individual status. The data collection spans six time steps: at 0, 4, 12, 18, 24, and 36 months. This approach is standard in medical treatment, although for optimal model performance, an equal distribution of time intervals would be preferable.

\subsubsection{Data Collection}
Data were collected from patient records in hospitals, and the following information was extracted for analysis: patient age, gender, distance to clinic, ratio of AIT cost to family income, allergen test results, etc., as well as patient VAS system score and medication score information, including baseline data of patients before injection therapy, adverse reactions to SCIT. For the descriptive analysis, categorical variables were given as numbers and percentages, and continuous variables were 
presented using mean, standard deviation, median, interquartile range (IQR), and minimum and maximum values. To address missing values, we tracked every patient, which allowed us to ensure the dataset's completeness. We did not remove outliers, aiming to follow real-clinical scenarios as closely as possible.

\subsubsection{Survey methods}
Adherence was defined as the accomplishment of three years of AIT. Non-adherence was defined as discontinuation of AIT at random time points during three years. The follow-up contents included (1) the main reasons for patients' discontinuation of treatment; (2) the duration of discontinuation of treatment, and (3) Allergic symptoms after discontinuation of treatment.

\subsection{Sequential models}

The focus of our study is the development of sequential models that can efficiently and accurately predict the progression of symptoms and adherence in patients undergoing SCIT. This involves a comprehensive analysis of the data collected, structured to provide insights into the treatment's effectiveness and patient compliance over time. Additionally, we explore and compare two distinct sequential models.

\subsubsection{Data}

\begin{table}[h]
\centering
\begin{tabular}{llccc}
\toprule
\multicolumn{2}{c}{\multirow{2}{*}{variables}} & \multicolumn{3}{c}{patients} \\ [2ex ]
\multicolumn{2}{c}{} & total & adherent & non-adherent  \\ 
\midrule
age & $\leq 12$ & 96 (46.7) & 40 & 56 \\
 & 13--17 & 30 (14.6) & 10 & 20 \\
 & $\geq 18$ & 79 (38.7) & 23 & 56 \\[1.2ex]
gender & Female & 62 (30.2) & 22 & 40 \\
 & Male & 143 (69.8) & 51 & 92 \\[1.2ex]
distance to clinic(km) & $\leq 10$ & 136 (66.3) & 56 & 80 \\
 & $> 10$ & 69 (33.7) & 17 & 52 \\[1.2ex]
cost/family income(\%) & $< 30$ & 107 (52.4) & 37 & 70 \\
 & 30--50 & 77 (37.4) & 32 & 45 \\
 & $> 50$ & 21 (10.2) & 4 & 17 \\[2ex]
EOS($\times$ 10\(^{9}\)/L)  & & 0.37; 0.41 & 0.36; 0.52 & 0.38; 0.36 \\
EOS \%  & & 0.05; 0.04 & 0.05; 0.05 & 0.05; 0.05 \\
$\Delta$NR(\%) & & 16.67; 59.70 & 30.00; 92.80 & 14.80; 50.00 \\
$\Delta$PNIF(\%) & & 11.90; 34.50 & 12.70; 39.30 & 11.10; 28.80 \\
total IgE (kU/L) & & 286; 543 & 340; 487 & 226; 555 \\
sIgE of Der p (kU/L) & & 30.80; 68.480 & 31.30; 74.40 & 30.40; 67.80 \\
sIgE of Der f (kU/L) & & 40.00; 68.20 & 40.60; 75.10 & 37.10; 65.70 \\
Der p SPT SI  & & 1.04; 0.58 & 1.00; 0.59 & 0.82; 0.55 \\
Der f SPT SI  & & 1.00; 0.50 & 0.82; 0.51 & 0.80; 0.45 \\ \bottomrule
\end{tabular}
\caption{Demographic and clinical data of the patients under subcutaneous immunotherapy. In the rows from Age to Cost/Family income, values indicate the number of patients (percentage, if available). Other rows represent the median and IQR. P-values are omitted due to their large values.}
\label{table:baseline}
\end{table}

\begin{table}[ht]
\centering
\begin{tabular}{lccccr}
\toprule
\multirow{2}{*}{\parbox{4cm}{\centering reasons for \\ SCIT withdrawal}} & \multicolumn{4}{c}{number of non-adherent patients} & \\ 
\cmidrule(lr){2-6}
& 5--12 mths & 13--18 mths & 19--24 mths & 25--36 mths & total by reason  \\ 
\midrule
no clinical improvement & 18 & 11 & 8 & 21 & 58 \\ 
medical issue & 3 & 1 & 2 & 0 & 6 \\ 
improved efficacy & 0 & 0 & 0 & 24 & 24 \\ 
schooling & 3 & 3 & 0 & 5 & 11 \\ 
side effects & 2 & 1 & 1 & 2 & 6 \\ 
COVID-19 & 9 & 7 & 3 & 1 & 20 \\ 
personal issue & 0 &  3& 0 & 4 & 7 \\ 
total by time period & 35 & 26 & 14 & 57 & 132 \\ 
\bottomrule
\end{tabular}
\caption{Detailed reasons for withdrawal from SCIT at different time points.}
\label{table:scit_withdrawal}
\end{table}

\begin{figure}[ht!]
    \centering
    \includegraphics[width=1\textwidth]{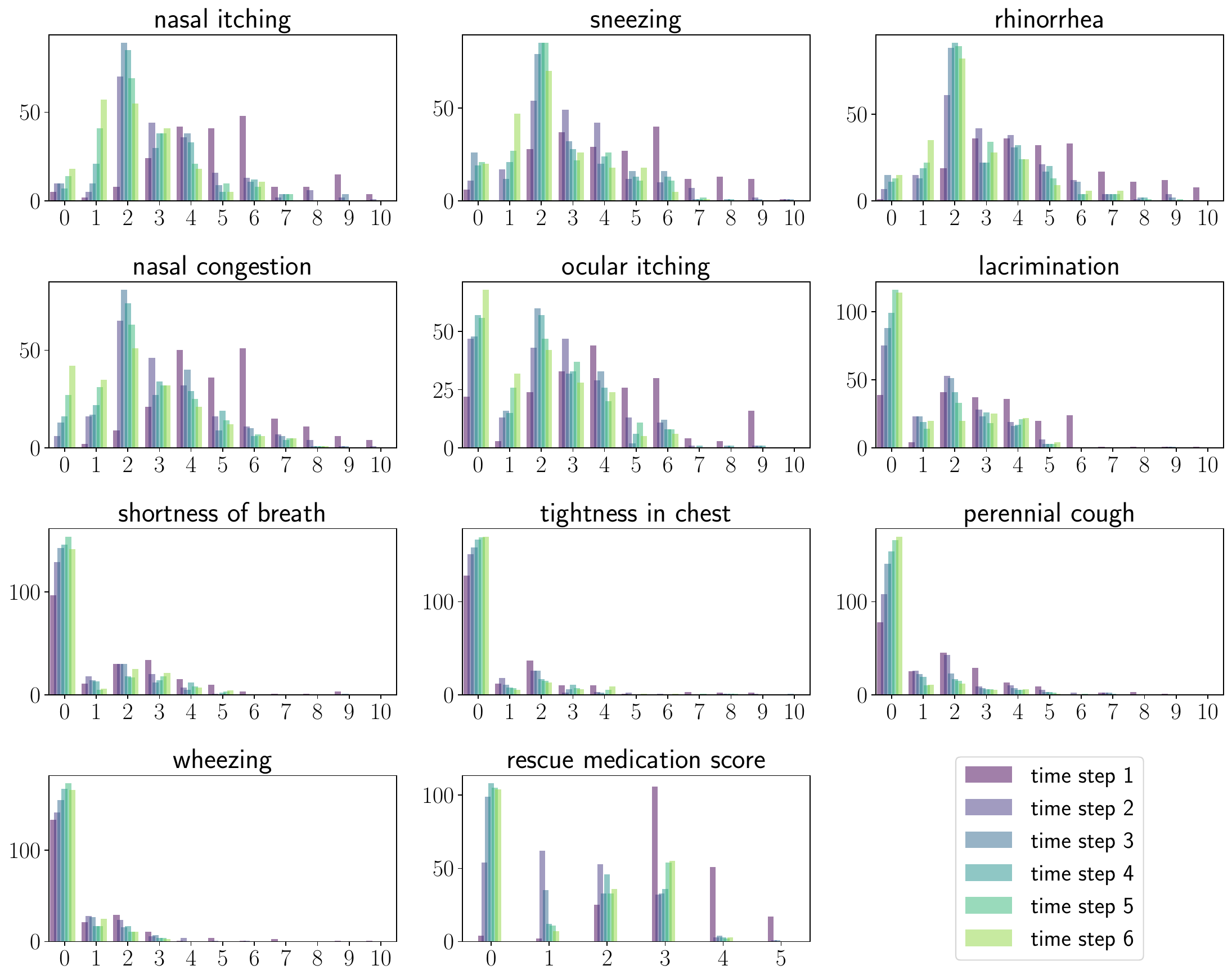}
    \caption{Histogram of scores across six time steps. Score value (horizontal axis) vs.\ count (vertical axis).
}
    \label{fig:histogram}
\end{figure}

We have a dataset $D$, comprising sequences $x_1, \dots, x_T \in \mathbb{R}^{11}$, $y_1, \dots, y_{T-1} \in \mathbb{R}^{1}$, and a corresponding action $a_t \in R^1$. In the context of healthcare, the observations encompass $y_t$ (see Table \ref{table:scit_withdrawal}) whether the patient will cease the treatment in the interval between the scoring measurements at $x_t$ and $x_{t+1}$.
The actions $a_t$ represent the ongoing medical procedures for the patient during the period from $x_{t}$ to $x_{t+1}$ (see Fig.~\ref{fig:histogram}). 
In this context, $a_t$ is binary, reflecting whether treatment is given, and is numerically equivalent to the adherence variable $y_t$. Despite their numerical equivalence, we maintain a distinction between action $a_t$ and adherence $y_t$ to enhance model clarity and accommodate future research expansions, potentially allowing for a wider range of action values.
For each patient, we possess basic information $s \in \mathbb{R}^{14}$ which includes age, gender, commute distance to clinic, ratio of cost to family income, eosinophils count, eosinophils percentage, nasal
allergen provocation test (change of nasal resistance, $\Delta \text{NR}(\% )$), peak nasal inspiratory flow.
$\Delta \mathrm{PNIF}(\%)$), serum total IgE level, sIgE of Dermatophagoides pteronyssinus (Derp), sIgE of Dermatophagoides farinae (Derf), skin prick test (Derp, Derf) (see Table \ref{table:baseline} for more details).

\subsubsection{Sequential latent variable model}

In our research, we use the Stochastic Latent Actor-Critic (SLAC) model \citep{lee2020stochastic}. Our application differs from the original use of SLAC which is typically associated with reinforcement learning. Instead, we use its sequential latent-variable model (SLVM) without Actor-Critic. This approach aligns with similar methodologies found in other works
\citep{krishnan2015deep,karl2016deep,gregor2018temporal}.
The choice of the SLAC model was motivated by its ability to facilitate more efficient learning and superior generalization in intricate environments.

The SLVM is fundamentally a framework that processes information in a step-by-step manner, capturing the dynamics of an environment or process over time. It constructs a latent representation of the data that it identifies and uses underlying patterns or structures within the dataset that probably are not immediately obvious. This capability makes it exceptionally suitable for tasks where understanding temporal relationships is crucial, such as predicting patient adherence in allergen immunotherapy. The model operates by generating a sequence of predictions, each informed by the data received up to that point, thereby enabling it to adapt and refine its understanding as more information becomes available. This methodological choice allows our research to use the strengths of SLAC in a novel context, applying it to the predictive modeling of patient behaviors in a healthcare setting.

After training the model, given the historical data up to step $t-1$, 
the model is capable of generating a patient's next scores from time step $t$ to $T$ directly from the latent space 
i.e., $p(x_{t:T} \mid x_{0:t-1}, a_{0:t-1} )$.
Similarly, it can predict the adherence of step $t:T-1$ at time step $t$, $p(y_{t:T-1} \mid x_{0:t}, a_{0:t-1} )$. See details of the model and the training process in Appendix~\ref{app:slac}.

\subsubsection{LSTM}
\label{sec:lstm}

As an alternative, Long short-term memory (LSTM) is a classical sequential neural-network model \citep{hochreiter1997long}. Given the historical data, in our implementation, an LSTM predicts the score $x_{t+1}$ and adherence $y_{t}$ in parallel.
The LSTM's autoregressive feature allows us to iteratively input its current predictions to predict subsequent outcomes, covering prediction from $x_{t+1}$ to $x_T$ and $y_{t}$ to $y_{T-1}$. 
Implementation details and the reasons for the model choice can be found in Appendix~\ref{app:lstm}.

\section{Results}

A total of 205 patients were enrolled in this study. The mean age was \(17.57 \pm 11.68\) years, children and adolescents represented the major population (61.3\,\%) in AIT treatment. Males (70\,\%) were predominantly represented. The population with a commute distance to the clinic within \(10\,\text{km}\) was 66 percent. Due to a great portion of juveniles from the cohort, the ratio of cost to family income instead of personal income was evaluated. The patients who undertook AIT cost less than 30\% of monthly family income and account for half the distribution of the population, while 12.9\% non-adherent patients undertook the 50\% financial burden in AIT treatment.

The change of nasal resistance (NR) and peak nasal inspiratory flow (PNIF) after nasal allergen provocation (NPT) was used to evaluate the severity of symptoms by combining the symptom score. The change of NR after NPT from the adherent group was higher than the non-adherent group \( (30\,\% \text{ vs } 14.8\,\%) \). The laboratory tests such as total IgE, sIgE of Der p and Der f, and SPT did not exhibit a significant difference between the two groups. For detailed characteristics of patients see Table \ref{table:baseline}.

The observed total non-adherence rate at the end of three years was 35.4\% and the median of the SCIT duration was 18 months in the study. The rate of dropout in the third year (43.0\%) was highest in comparison to the end of the first year (26.5\%) and the second year (30.0\%). The reason for the withdrawal from the patients included the concern of COVID-19, especially at the beginning of 2020 accounting for a 25\% portion of the non-adherent patients in the first year. The most influential reason for the withdrawal was unreached expectations for clinical improvement (43.9\%). Medical issues including pregnancy status during the treatment period and other physical disorders were collected from patients leading to the withdrawal of SCIT (4.5\%). The significantly improved symptoms contributed to the reason for dropout, especially after two years SCIT treatments (18.2\%). The recorded cases from side effects accounted for 4.5\%, comprising the local and systematic adverse reactions (see Table~\ref{table:scit_withdrawal}).

We have a total of 205 samples, which we have randomly divided into a test dataset comprising 20\%, i.e., 41 samples. For our analysis, we employ a five-fold cross-validation approach. Additionally, we apply zero-mean and unit standard deviation (STD) normalization to the variables $x$ and $s$.

The Root Mean Square Error (RMSE) metric is used to evaluate the precision of our medical score predictions. Furthermore, to assess the adherence predictions, we use a comprehensive set of metrics including accuracy, precision, recall, and the F1 score, each offering a unique perspective on the performance of our predictive models. Most of the figures in this study are presented using boxplots.

In all results in this study, the uncertainties for both models are calculated using \textit{five-fold cross-validation}. In addition, as SLAC is a probabilistic model, we also perform 100 samples from the latent space to compute its uncertainty.

\subsection{One-Step Prediction}

\begin{figure}[ht!]
    \centering
    \includegraphics[width=0.5\textwidth]{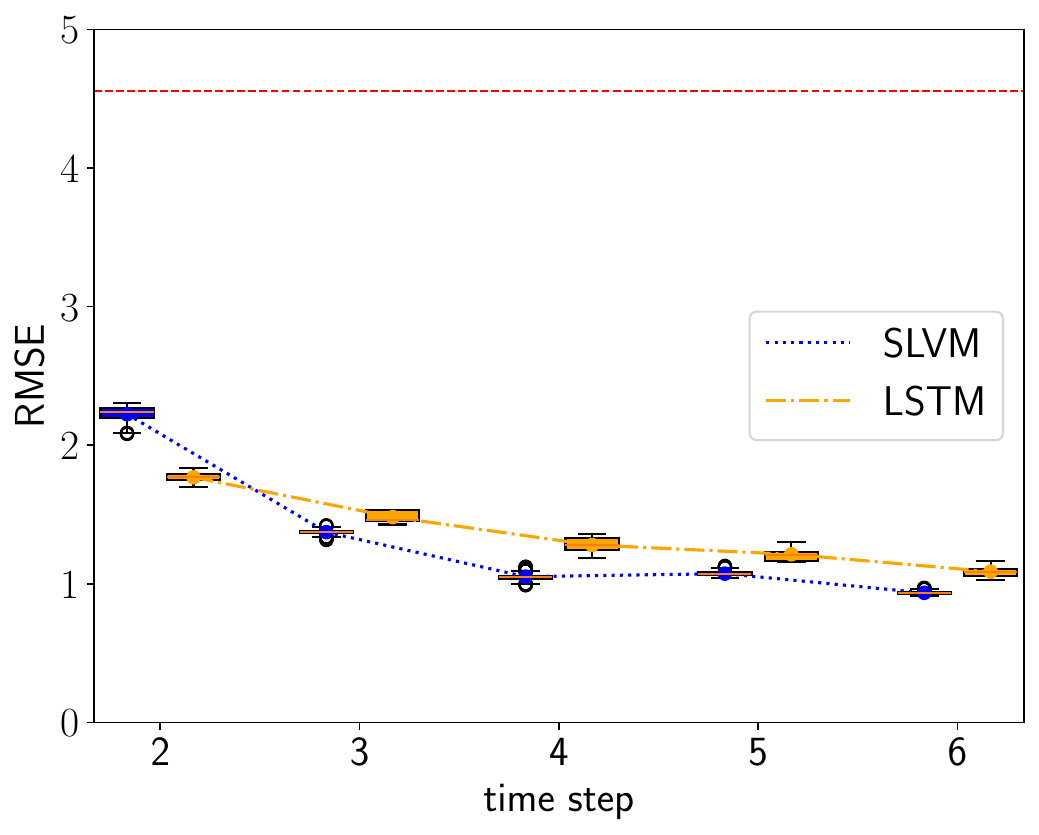}
    \caption{RMSE of the prediction step by step. The red dashed line is the RMSE of random prediction with Uniform distribution.  See Fig.~\ref{fig:all_feature_onestep_slac} and \ref{fig:all_feature_onestep_lstm} for more details.
}
    \label{fig:rmse_onestep}
\end{figure}

\begin{figure}[ht!]
    \centering
    \includegraphics[width=0.5\textwidth]{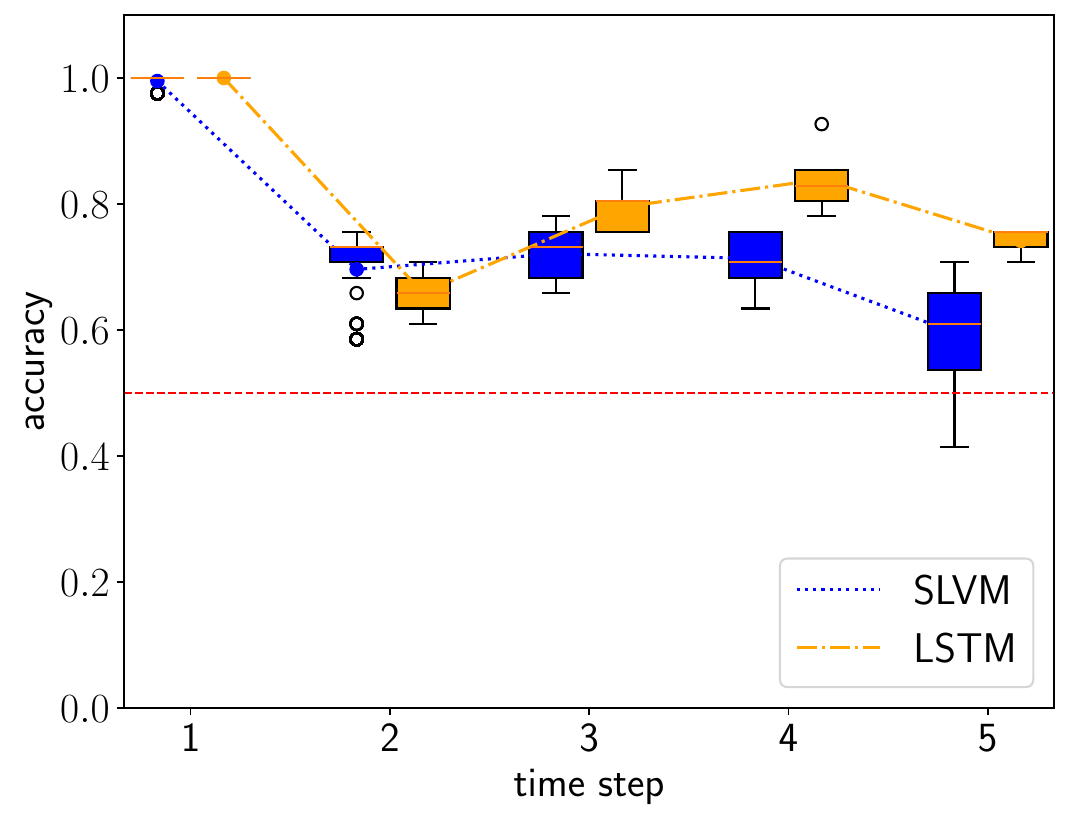}
    \caption{Accuray of the prediction step by step. The red dashed line is the accuracy of random prediction with Uniform distribution. See Table \ref{table:classification} for more details.
}
    \label{fig:acc_onestep}
\end{figure}

\begin{figure}[ht!]
    \centering
    \includegraphics[width=1.0\textwidth]{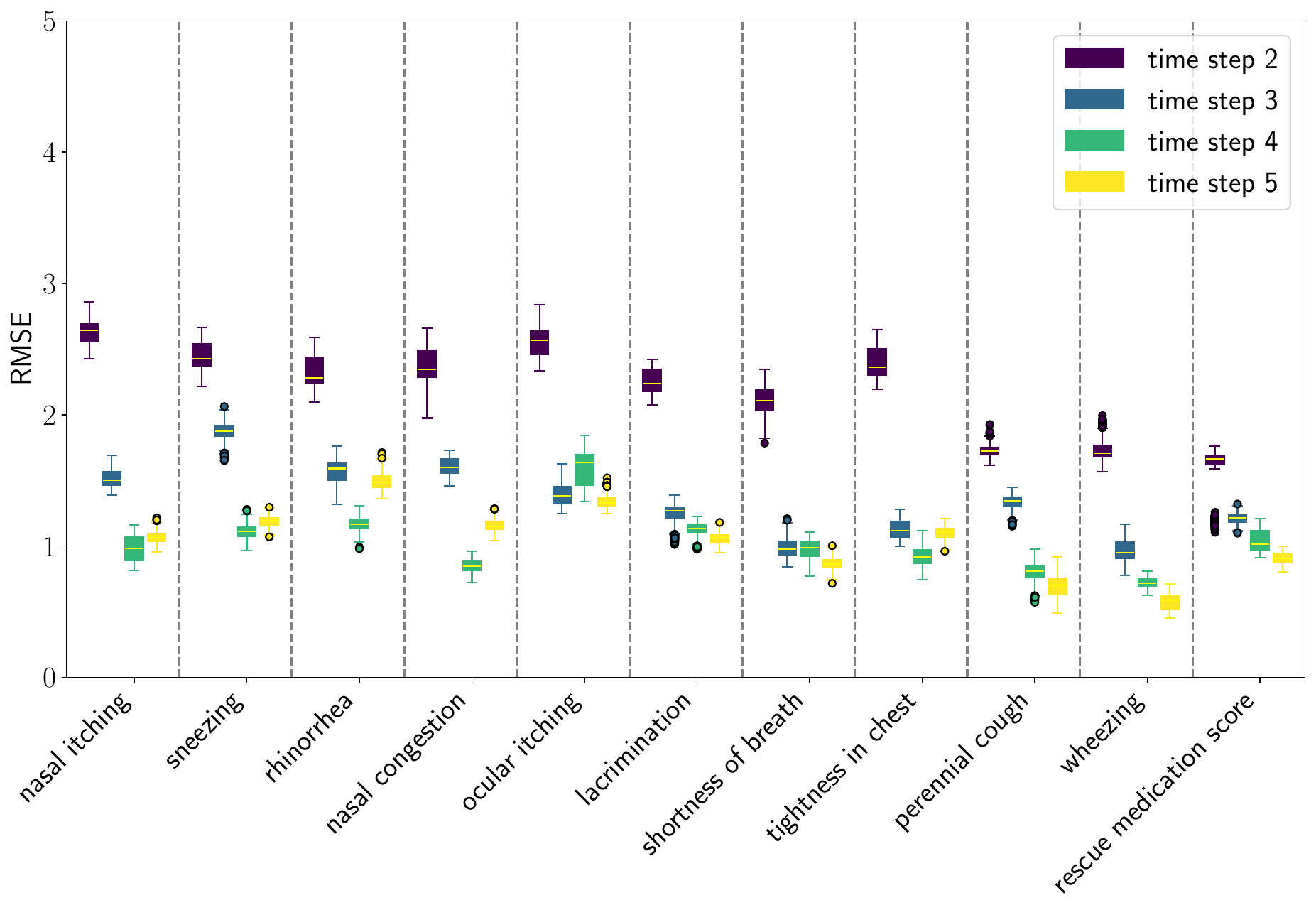}
    \caption{RMSE of SLAC one-step prediction across various scores and time steps. 
}
    \label{fig:all_feature_onestep_slac}
\end{figure}

\begin{figure}[ht!]
    \centering
    \includegraphics[width=1.0\textwidth]{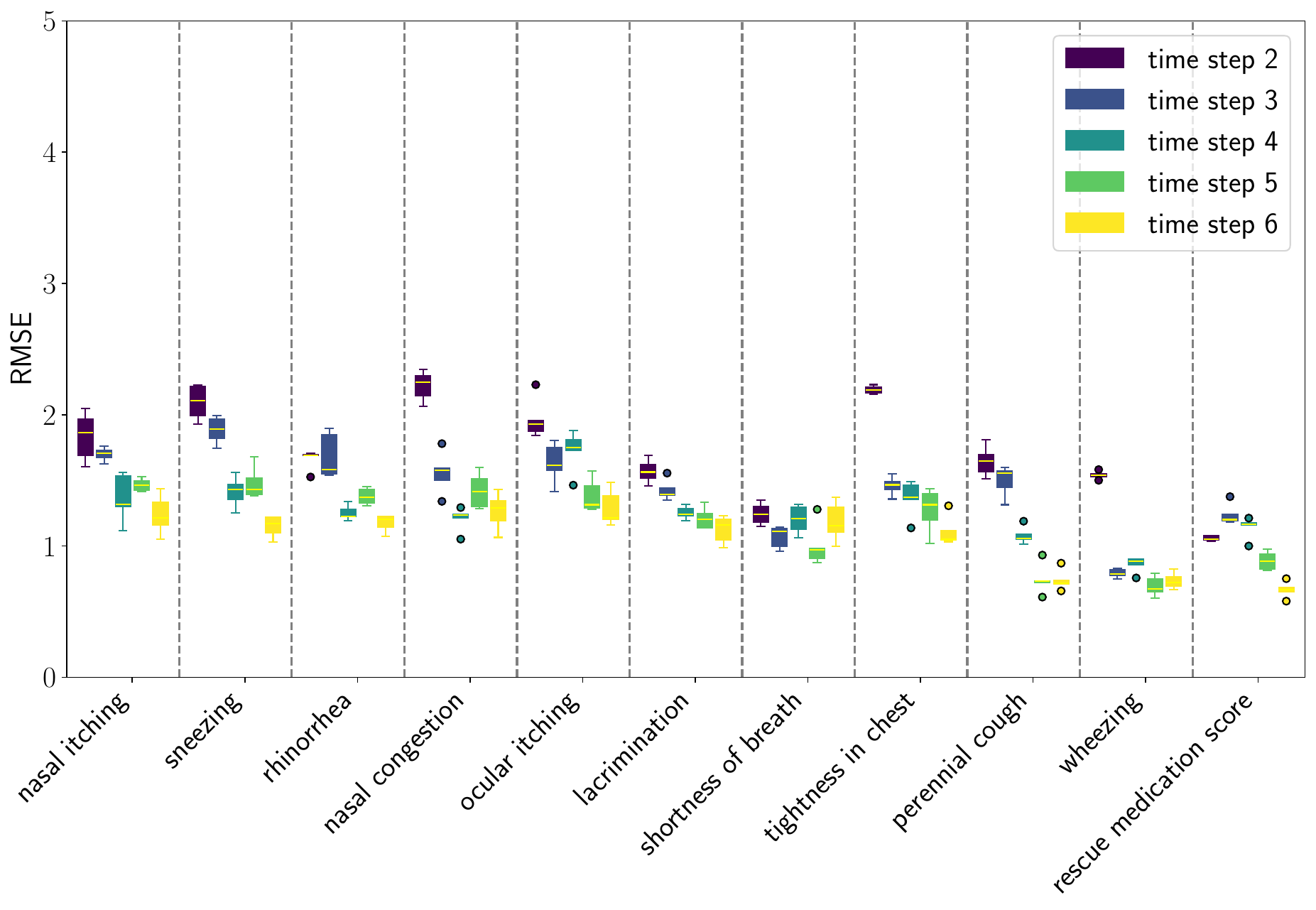}
    \caption{RMSE of LSTM one-step prediction across various scores and time steps.
}
    \label{fig:all_feature_onestep_lstm}
\end{figure}

\begin{table}[ht!]
\centering
\begin{tabular}{llccccr}
\toprule
metric & model & time step 1 & time step 2 & time step 3 & time step 4 & time step 5 \\
\midrule
accuracy & LSTM & $\mathbf{1.00 \pm 0.00}$ & $0.66 \pm 0.03$ & $\mathbf{0.80 \pm 0.04}$ & $\mathbf{0.84 \pm 0.05}$ & $\mathbf{0.74 \pm 0.02}$ \\
 & SLVM & $1.00 \pm 0.01$ & $\mathbf{0.70 \pm 0.06}$ & $0.72 \pm 0.04$ & $0.71 \pm 0.04$ & $0.60 \pm 0.06$ \\
\addlinespace
precision & LSTM & $\mathbf{1.00 \pm 0.00}$ & $0.72 \pm 0.01$ & $\mathbf{0.86 \pm 0.06}$ & $\mathbf{0.90 \pm 0.05}$ & $\mathbf{0.62 \pm 0.03}$ \\
 & SLVM & $\mathbf{1.00 \pm 0.00}$ & $\mathbf{0.75 \pm 0.03}$ & $0.74 \pm 0.03$ & $0.71 \pm 0.03$ & $0.44 \pm 0.05$ \\
\addlinespace
recall & LSTM & $\mathbf{1.00 \pm 0.00}$ & $0.86 \pm 0.06$ & $0.83 \pm 0.05$ & $0.82 \pm 0.08$ & $0.61 \pm 0.03$ \\
 & SLVM & $1.00 \pm 0.01$ & $\mathbf{0.87 \pm 0.06}$ & $\mathbf{0.90 \pm 0.03}$ & $\mathbf{0.86 \pm 0.04}$ & $\mathbf{0.70 \pm 0.10}$ \\
\addlinespace
F1 score & LSTM & $\mathbf{1.00 \pm 0.00}$ & $0.79 \pm 0.03$ & $\mathbf{0.84 \pm 0.03}$ & $\mathbf{0.85 \pm 0.05}$ & $\mathbf{0.62 \pm 0.03}$ \\
 & SLVM & $1.00 \pm 0.00$ & $\mathbf{0.81 \pm 0.04}$ & $0.81 \pm 0.02$ & $0.78 \pm 0.03$ & $0.54 \pm 0.06$ \\
\bottomrule
\end{tabular}
\caption{Comparison of LSTM and SLAC over different time steps. The results are expressed as a mean ± standard deviation. The better results are highlighted in bold.}
\label{table:classification}
\end{table}

In this experiment, our focus is on predicting the immediate next step. Within the SLVM, the prediction of $y_{t}$ is based on the sequence $x_{1:t}$ and actions $a_{1:t-1}$. Additionally, we forecast the subsequent state $x_{t+1}$ using the sequence $x_{1:t}$ along with actions $a_{1:t}$. In contrast, for LSTM, the predictions for both $y_{t}$ and the next state $x_{t+1}$ are derived from $x_{1:t}$ and $y_{1:t-1}$.

As illustrated in Fig.~\ref{fig:rmse_onestep}, SLVM surpasses LSTM in performance beginning at time step two. The figure indicates that with an increased amount of historical data (additional time steps), SLVM achieves greater RMSE. Both SLAC and LSTM demonstrate considerably better over random prediction methods. Further insights are provided in Fig.~\ref{fig:all_feature_onestep_slac} and Fig.~\ref{fig:all_feature_onestep_lstm}, which provides detailed representations of each feature. In the prediction of specific local symptoms score, SLVM performs an improvement in error after step two with all parameters compared to LSTM. The results from RMSE in nasal and ocular symptoms display relatively high values compared to those for lower respiratory tract symptoms. This can be attributed to the fact that the majority of patients in the cohort predominantly exhibited nasal and ocular symptoms, which presented a wide range of scores.

Fig.~\ref{fig:acc_onestep} demonstrates that from steps two to four, accuracy in adherence predictions improves with the inclusion of additional information. The first step shows a notable bias, as it only includes data from adherent patients, as detailed in Sec.~\ref{sec:method_treatment}. Nonetheless, both models adeptly manage this bias and achieve high-accuracy predictions. Prediction for the sixth step is not conducted due to the cessation of treatment by the hospital. In the fifth step, there is a decline in accuracy, likely due to the extended time interval of 12 months. In future research, it would be worthwhile to explore whether adopting a consistent interval for data collection could enhance the outcomes of longitudinal prediction.
Table \ref{table:classification} illustrates details of the classification for one-step prediction.
Initially, both models exhibit perfect performance in Accuracy, Precision, and Recall at the first time step, but diverge in subsequent steps.
In terms of Accuracy, LSTM generally outperforms SLVM, particularly evident at time steps three, four, and five. For Precision, LSTM again shows superior performance in the later time steps, except at time step two where SLVM marginally leads. However, in the Recall metric, SLVM surpasses LSTM from time step two onwards, indicating its strength in correctly identifying positive cases.
The F1 score, which balances precision and recall, shows LSTM generally ahead, except at time step two where SLAC has a slight edge. This metric indicates LSTM's balanced capability in both precision and recall, especially in the later time steps.
Overall, while both models start equally strong, LSTM demonstrates greater consistency and effectiveness across most metrics and time steps. SLVM, while lagging slightly behind in accuracy and precision, shows its robustness in recall, especially in the middle to later time steps.

\subsection{Rollouts}

\begin{figure}[ht!]
    \centering
    \includegraphics[width=1\textwidth]{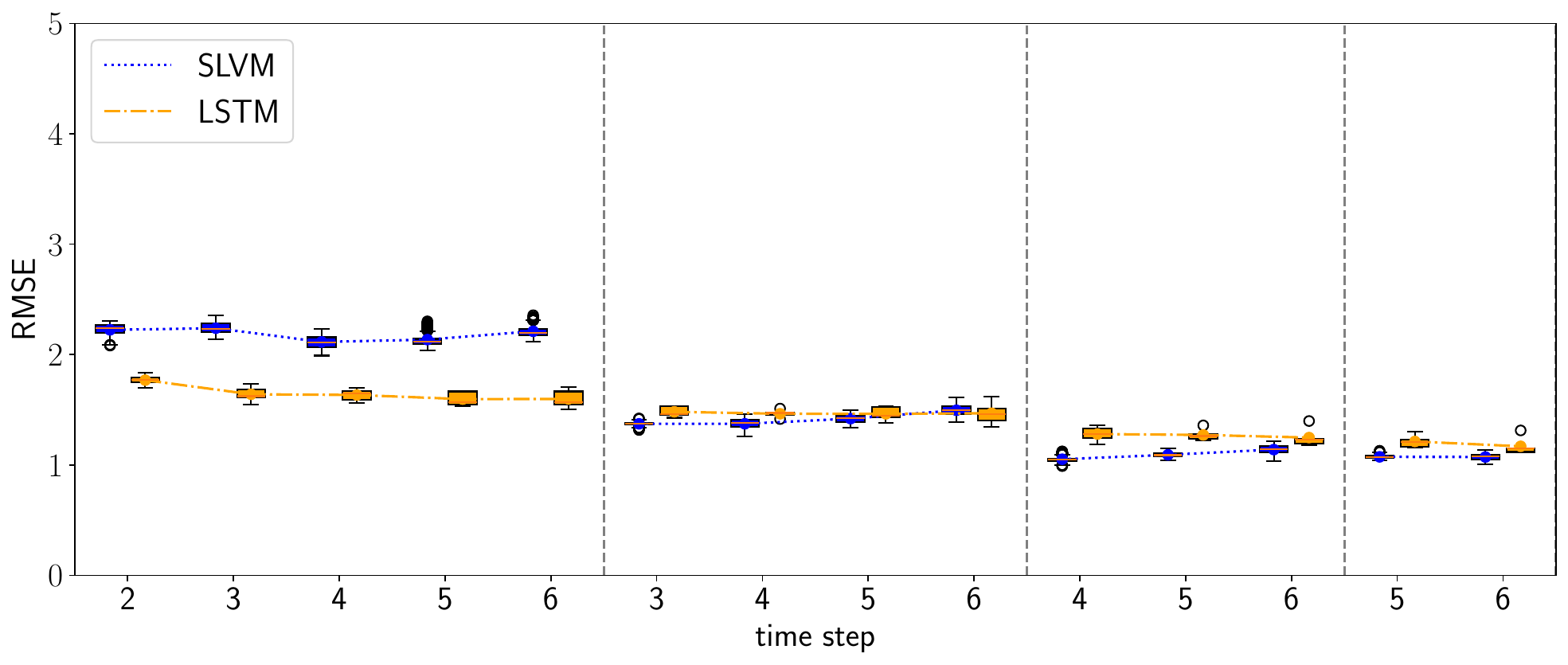}
    \caption{RMSE of the rollout prediction. The first time step in each subplot represents the beginning of the rollout time step.
}
    \label{fig:rmse_rollout}
\end{figure}

\begin{figure}[ht!]
    \centering
    \includegraphics[width=1\textwidth]{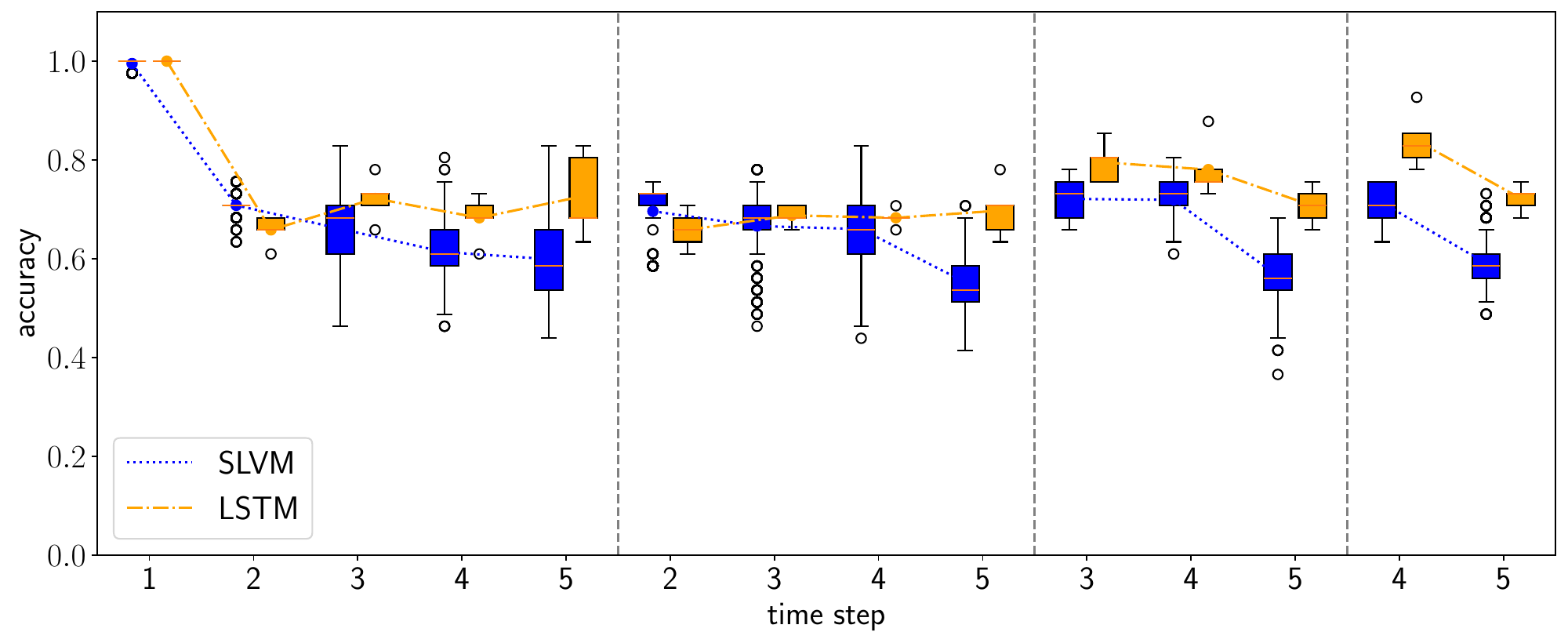}
    \caption{Accuracy of the rollout prediction. The first time step in each subplot represents the beginning of the rollout time step.
}
    \label{fig:accuracy_rollout}
\end{figure}

In the rollout experiment, our focus extends to a longer-term prediction. The SLAC prediction of $y_{t:T-1}$ and $x_{t+1:T}$ are computed based on $x_{1:t}$ and $a_{1:t-1}$. Actions, $\{a_i:i\geq t\}$ are inferred from the model's output, $y_t$. Moreover, for time steps greater than $t$, we employ a prior in the latent space, which eliminates the need for the input of $x_{t:T}$.
In the LSTM model, the predictions for $y_{t:T-1}$ and $x_{t+1:T}$ are based on $x_{1:t}$ and $y_{1:t-1}$.

Fig.~\ref{fig:rmse_rollout} and \ref{fig:accuracy_rollout} illustrate the performance of our model in multi-step predictions. Similar to one-step predictions, the accuracy generally improves with the availability of more information, except in the case of the adherence prediction at the fifth step. The results demonstrate the model's proficiency in making long-term predictions.

\subsection{Model as a simulator}

Given the initial condition of a patient, we can assess the outcomes of various interventions. Clinically, if the patient's adherence to treatment significantly impacts the prognosis (and there is a possibility of non-adherence), it becomes imperative for the doctor to emphasize treatment compliance. Conversely, if adherence makes little difference, it suggests the therapeutic approach may be ineffective for this patient, allowing the doctors to emphasize adherence efforts.

To evaluate the impact of varying actions on SLAC's performance, we analyze how different actions affect the resulting scores. In the absence of a ground truth with diverse actions for the same patient, our focus shifts to examining whether the states are responsive to changes in actions. Considering initial states $x_{1:3}$ and actions $a_{1:2}$, we do rollouts with $a_{3:5}$, alternating between one and zero. This controlled alteration reveals that the average predicted value of $x_6$ under these conditions is $-0.20$. This value is computed from the prediction outcomes for actions with ones minus those for actions with zeros. 
The result indicates that our model can be used as a simulator for doctors to see the impact of different treatments/therapies.
Since the LSTM does not have similar functions (see Sec.~\ref{sec:lstm}), we only show the SLAC results.

\subsection{Interpretability}

\begin{figure*}[ht!]
    \centering
    \includegraphics[width=1.0\textwidth]{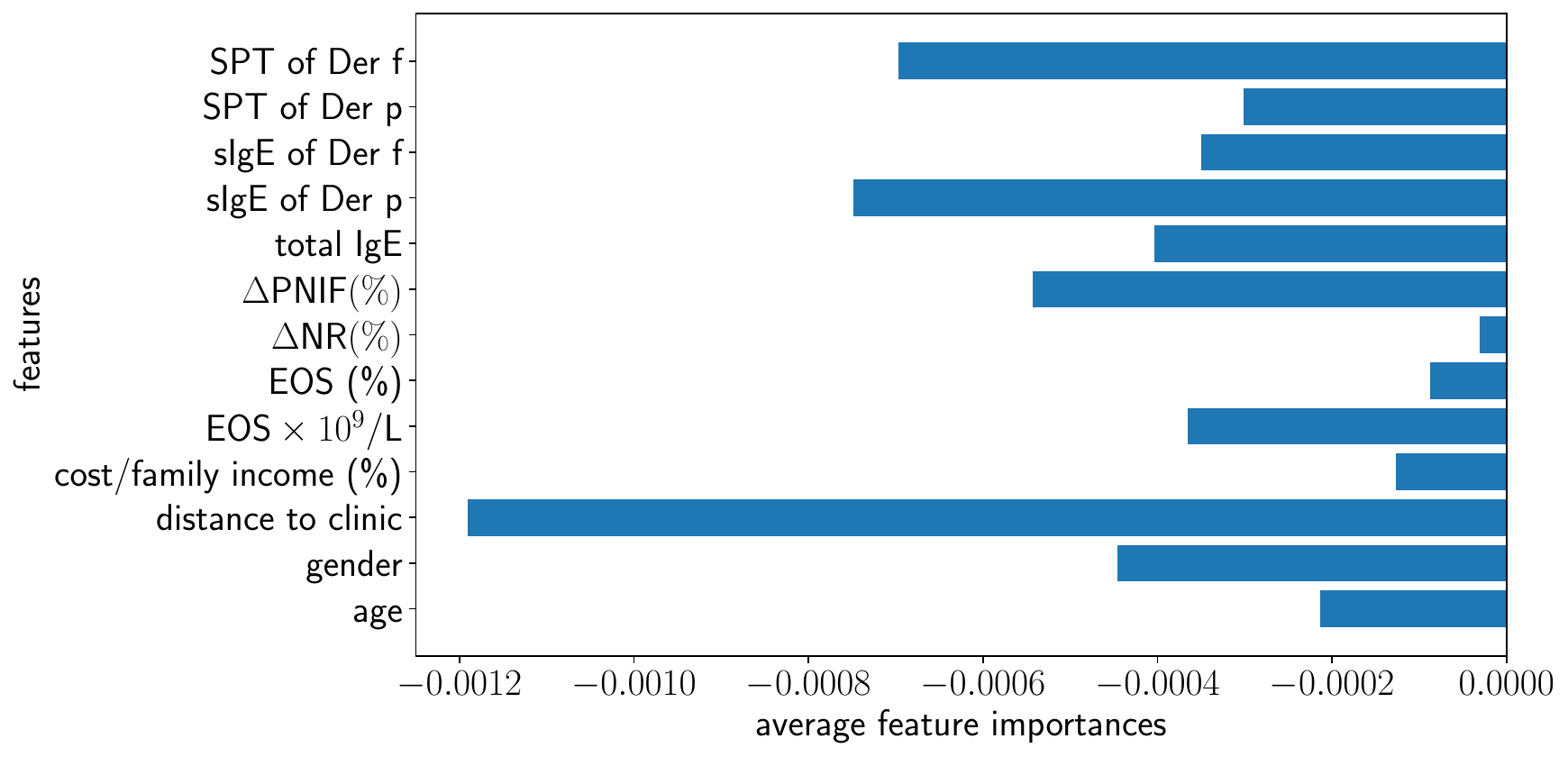}
    \caption{Importances of the factors. 
}
    \label{fig:importance_slac}
\end{figure*}

Previous models and methods have been developed for interpreting machine learning algorithms, including SHAP \citep{lundberg2017unified} and Captum \citep{kokhlikyan2020captum}. We opt for Captum, as it integrates more seamlessly with PyTorch-based code. 
We perform the measure of the factor importance using Integrated Gradients of Captum for SLAC (see Fig.~\ref{fig:importance_slac}).
The magnitude of features highlights the significance of the model's prediction for a specific class.
The distance to the clinic significantly impacts patient adherence, especially if a patient is located far from the clinic or has relocated, as they are more likely to discontinue their visits. Following the distance, SPT of Der f and sIgE of Der f greatly influence the adherence. In contrast,$\Delta \text{NR}(\%)$, EOS$(\%)$, and the cost/family income$(\%)$ have minimal impacts.

\section{Discussion and Conclusion}

  The reported adherence rates of SCIT ranged from around 23\% to 90\%, due to the non-uniform follow-up duration (2--4 years) \citep{passalacqua2013adherence,lee2019factors,lemberg2017sublingual,yang2018risk}. Poor adherence in the three to five-year time span of AIT is an obstacle to reaching allergen tolerance and symptom remission. Recently "adherence and persistence in AIT (APAIT)" checklist was proposed to assists researchers in assessing adherence or persistence to AIT treatment\citep{pfaar2023adherence}. The present study is the first research regarding the application of machine learning models in the adherence prediction of SCIT in AR patients. From our study, the accomplishment rate of the three-year treatment cycle was relatively low (35.4\%), while the dropout rate after two years accounts for half (42.8\%) in the whole non-adherence cohort. Several researchers focus on these variables impacting adherence to medical behaviors to enhance the intervention approach to reduce the withdrawal caused by disease-unrelated reasons. Even though the COVID-19 pandemic affected the adherence of the majority of the patients in the three-year cycle, the first year since the pandemic's outbreak appears to have fundamentally built a barrier to patients, similar to the finding from \citet{liu2021compliance} that 11\% dropouts in the two years SCIT was observed caused by COVID-19. We excluded the patients who dropped out in the dose buildup phase within four months to minimize the dose-origin impact. Due to the uncovered cost from the public health care system and commercial insurance, financial burden accounts for a non-negligible factor in influencing the patient's decision-making. A similar finding from \citet{lourencco2020subcutaneous} indicated that economic reasons contributed to the most frequent cause of SCIT cessation.

Recently, as the presented studies focusing on the medication adherence prediction of non-communicable diseases such as diabetes, hypertension, cancer, and chronic respiratory diseases regarding machine learning models were introduced into the application, the systematic monitoring of patients' adherence behaviors remarkably re-tailored the disease management and enhanced medical decision-making \citep{kanyongo2023machine}
Due to the multidimensional variables in the prediction of adherence collected for analysis, machine learning models exhibit advantages in automatic feature selection, interaction effects, scalability, robustness, and so on compared to traditional regression analysis.

 \cite{mousavi2022determining}  demonstrated the effectiveness of a hybrid model that combines neural networks and genetic algorithms for predicting diet adherence. \cite{wang2020applying} explored another hybrid model that integrates neural networks and support vector machines to predict nonadherence in Crohn's Disease patients by streamlining the intervention process in medicine-taking. Both methods have shown excellent performance.
The deployment of machine-learning algorithms in the prediction of adherence in cardiovascular disease including random forests, support vector machines, and neural networks showed the accuracy ranged from 0.53 to 0.97~\citep{zakeri2022application,mirzadeh2022use}.
The ensemble learning model in the prediction of adherence from the patients who conducted self-administer injections proposed by \citet{gu2021predicting} achieved a good performance and generalization properties based on the fusion of multiple heterogeneous classifiers. 
In the field of allergen immunotherapy, \citet{yao2023predicting} introduced a machine-learning model with an improved DFSSA algorithm to predict the therapeutic efficacy of AIT for asthma using clinical characteristics and serum allergen detection metrics. 
However, these non-sequential methods generally predict only the final outcome, neglecting the complexities of intermediate stages.
A sequential model that can make predictions at any specific time step would significantly enhance the ability for early intervention.
\citet{hsu2022medication} investigated the advantages of incorporating patient history into the prediction of medication adherence. They assessed the performance of temporal neural network models, particularly LSTM and simple recurrent 
neural networks, and compared these with non-temporal neural networks, ridge classifiers, and logistic regression. To optimize the efficacy of cognitive training for older adults, \citet{singh2022deep} employed multivariate time series analysis and developed personalized models for each patient to capture their unique adherence patterns. However, the sequential data of patients is often characterized by fluctuating adherence and high dropout rates, resulting in uneven, unaligned, and missing values in the time series data. To address this challenge, \citet{schleicher2023prediction} applied change point detection to identify phases with varying dropout rates, presented methods for handling uneven and misaligned time series, and used time series classification to predict the user's phase. These models, however, overlook the significance of score prediction in SCIT treatment. Our study advances this methodology by integrating a state-action model capable of predicting both adherence and score/state. This enhancement facilitates a more accurate and comprehensive whole-process management of AR patients in SCIT treatment.

The proposed prediction models can help clinicians dynamically measure the effectiveness of adherence interventions including more frequent reminders or engagement strategies, such that healthcare teams can focus on these individuals and proactively provide them with additional support. Moreover, in the surveillance of local symptom scores and rescue medicine up-take, the models offer a clinical evidence-based approach to precisely predict the risk of non-adherence in patient-centered care precisely. Especially for the potential risk of withdrawal caused by medical issues or related side effects, our models suggest multidimensional observational parameters for timely offering medical intervention and increasing patient engagement by participating in shared decision-making. The implementation of integration of the models with existing healthcare workflows is challenging, while the application of telesystem and online consultation would improve the work efficacy in immunotherapy centers and facilitate patient's self-management.

Our study demonstrates notable findings in the domain of patient adherence prediction in subcutaneous immunotherapy. The comparison between the SLAC model and LSTM model reveals the distinct strengths and limitations of each approach. Notably, SLAC exhibits greater flexibility, and it outperforms  LSTM in score prediction. This advantage likely stems from its ability to efficiently learn and generalize in complex environments. Conversely, the LSTM model shows better performance in predicting adherence, indicating its potential usage in scenarios. Both models demonstrate the capability to handle longer sequences, extending beyond one-step prediction. This ability is crucial in medical settings where long-term patient monitoring and prediction are essential for effective treatment planning.

Overall, the study underscores the importance of selecting the appropriate model based on the specific requirements of the task, whether it be flexibility, precision in score prediction, or adherence prediction. The findings contribute to the growing field of machine learning applications in healthcare, particularly in enhancing patient-centered treatment strategies through accurate and personalized predictions. Future research could focus on evaluating the SLAC model's performance in simulating various actions, further enriching its applicability in clinical settings. Additionally, the generalization to other diseases or the application of our models would be an interesting direction for future research.

\section*{Conflict of Interest Statement}

The authors declare that the research was conducted in the absence of any commercial or financial relationships that could be construed as a potential conflict of interest.

\section*{Data Availability Statement}
The dataset for this study can be found in a GitHub repository \url{https://github.com/leexxe/Subcutaneous-Immunotherapy-Dataset}.

\bibliographystyle{Frontiers-Harvard} %
\bibliography{reference}

\appendix

\section{Latent sequential variable model}
\label{app:slac}

\begin{figure}[!ht]
  \centering
  \begin{tikzpicture}[
  node distance=0.5mm, auto,
  empty/.style={shape=circle, minimum size=7mm, inner sep=0mm},
  obs/.style={shape=circle, draw=black, fill=black!20, minimum size=7mm, inner sep=0mm},
  unobs/.style={shape=circle, draw=black, minimum size=7mm, inner sep=0mm},
  gen/.style={->, draw=black, thick, -{Stealth[length=2mm]}},
  inf/.style={->, draw=black, densely dashed, bend right=22.5, -{Stealth[length=2mm]}},
]

  \node[empty] (z2_05)   []                 {};
  \node[unobs]   (z2_1)    [right=of z2_05]   {$z_{1}^2$};
  \node[empty] (z2_15)   [right=of z2_1]    {};
  \node[empty] (z2_2)    [right=of z2_15]   {$\cdots$};
  \node[empty] (z2_25)   [right=of z2_2]    {};
  \node[unobs]   (z2_t)    [right=of z2_25]   {$z_t^2$};
  \node[empty] (z2_tp05) [right=of z2_t]    {};
  \node[unobs]   (z2_tp1)  [right=of z2_tp05] {$z_{t+1}^2$};

  \node[unobs]   (z1_1)    [below=of z2_05]   {$z_1^1$};
  \node[empty] (z1_15)   [below=of z2_1]    {};
  \node[empty] (z1_2)    [below=of z2_15]   {$\cdots$};
  \node[empty] (z1_25)   [below=of z2_2]    {};
  \node[unobs]   (z1_t)    [below=of z2_25]   {$z_t^1$};
  \node[empty] (z1_tp05) [below=of z2_t]    {};
  \node[unobs]   (z1_tp1)  [below=of z2_tp05] {$z_{t+1}^1$};
  \node[empty] (z1_tp15) [below=of z2_tp1]  {};

  \node[obs]   (x_1)     [below=of z1_15]   {$x_1$};
  \node[obs]   (x_t)     [below=of z1_tp05] {$x_t$};
  \node[obs]   (x_tp1)   [below=of z1_tp15] {$x_{t+1}$};

  \node[obs]   (a_1)     [above=of z2_15]   {$a_{1}$};
  \node[empty] (a_2)     [above=of z2_25]   {};
  \node[obs]   (a_t)     [above=of z2_tp05] {$a_{t}$};
  \node[empty] (s0)   [left=of z1_1]                 {};
  \node[obs] (s) [below=of s0] {$s$};
  \node[empty] (z1_tp2) [right=of z1_tp15] {$\cdots$}; %
  \node[empty] (z2_tp15)   [right=of z2_tp1]    {};
  \node[empty] (z2_tp2)    [right=of z2_tp15]   {$\cdots$};

  \node[obs] (y_1) [below=of x_1] {$y_1$};
  \node[obs] (y_t) [below=of x_t] {$y_t$};
  \node[obs] (y_tp1) [below=of x_tp1] {$y_{t+1}$};

  \path (z2_1) edge [gen] (z2_2)
        (z2_2) edge [gen] (z2_t)
        (z2_t) edge [gen] (z2_tp1)

        (z2_1) edge [gen] (x_1)
        (z2_t) edge [gen] (x_t)
        (z2_tp1) edge [gen] (x_tp1)

        (z1_1) edge [gen] (z2_1)
        (z1_t) edge [gen] (z2_t)
        (z1_tp1) edge [gen] (z2_tp1)

        (z2_1) edge [gen] (z1_2)
        (z2_2) edge [gen] (z1_t)
        (z2_t) edge [gen] (z1_tp1)

        (z1_1) edge [gen] (x_1)
        (z1_t) edge [gen] (x_t)
        (z1_tp1) edge [gen] (x_tp1)

        (a_1) edge [gen] (z2_2)
        (a_1) edge [gen] (z1_2)
        (a_2) edge [gen] (z2_t)
        (a_2) edge [gen] (z1_t)
        (a_t) edge [gen] (z2_tp1)
        (a_t) edge [gen] (z1_tp1)

        (x_1) edge [inf, bend right=-22.5] (z1_1)
        (a_1) edge [inf] (z1_2)
        (z2_1) edge [inf] (z1_2)
        (x_t) edge [inf, bend right=-22.5] (z1_t)
        (a_2) edge [inf] (z1_t)
        (z2_2) edge [inf] (z1_t)
        (x_tp1) edge [inf, bend right=-22.5] (z1_tp1)
        (a_t) edge [inf] (z1_tp1)
        (z2_t) edge [inf] (z1_tp1)

        (s) edge [inf] (z1_1)  %

        (z2_tp1) edge [gen] (z2_tp2)
        (z2_tp1) edge [gen] (z1_tp2)
        (z2_tp1) edge [inf] (z1_tp2)

        (z1_1) edge [gen, bend left=-22.5] (y_1)
        (z1_t) edge [gen, bend left=-22.5] (y_t)
        (z1_tp1) edge [gen, bend left=-22.5] (y_tp1)
        (z2_1) edge [gen, bend right=-15.5] (y_1)
        (z2_t) edge [gen, bend right=-15.5] (y_t)
        (z2_tp1) edge [gen, bend right=-15.5] (y_tp1);
\end{tikzpicture}
  \caption{Schematic of the LSVM part of SLAC. Solid and dashed lines denote the generative and inference model pathways, respectively. 
  The gray circles represent observed data, and the white circles denote latent variables. The figure is adapted from \citep{lee2020stochastic}.}
  \label{fig:lvm}
\end{figure}
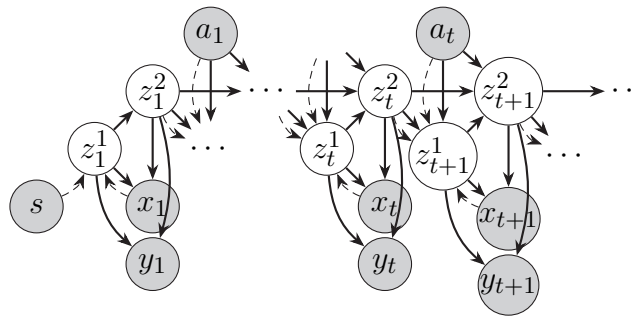

The sequential latent variable model (SLVM) of the SLAC consists of an inference model and a generative model (see Fig.~\ref{fig:lvm}).
The inference model in a sequential latent-variable model typically aims to approximate the posterior distribution of the latent variables given the observed data. It tries to infer the hidden states $z$ based on the observed inputs $x$ and initial states  
$s$. The inference models the probability distributions of the latent variables $z^1$ and $z^2$ at different time steps. $q_\phi$ denotes the variational distribution parameterized by $\phi$,
\begin{align}
    z^1_1 & \sim q_\phi (z_1^1 \mid x_1, s) \\
    z^2_1 & \sim p_\phi (z^2_1 \mid z_1^1) \\
    z^1_{t+1} & \sim q_\phi(z^1_{t+1} \mid x_{t+1}, z^2_t, a_t) \\
    z^2_{t+1} & \sim p_\phi (z^2_{t+1} \mid z^1_{t+1}, z^2_t, a_t).
\end{align}

The generative model, on the other hand, describes how the observed data is generated from the latent variables. The generative model is the probability distribution of both the initial latent states and their transitions over time, as well as the likelihood of the observations given the latent states, with $p_\phi$ indicating the parameterized generative distribution.
\begin{align}
    z^1_1 & \sim p (z_1^1) \\
    z^2_1 & \sim p_\phi (z^2_1 \mid z_1^1) \\
    z^1_{t+1} & \sim p_\phi(z^1_{t+1} \mid z^2_t, a_t) \\
    z^2_{t+1} & \sim p_\phi (z^2_{t+1} \mid z^1_{t+1}, z^2_t, a_t) \\
    x_t & \sim p_\phi(x_t \mid z_t^1, z^2_t) \\
    y_t & \sim p_\phi(y_t \mid z_t^1, z^2_t).
\end{align}

We have the evidence lower bound (ELBO):
\begin{align}
    \label{EqnELBO}
    \log p_\phi(x_{1:t+1} \vert a_{1:t}) \geq & \bigg[ \mathop{\mathbb{E}}_{(x_{1:T}, a_{1:T-1}) \sim D} \bigg[ \mathop{\mathbb{E}}_{z_{1:T} \sim q_\phi} \sum^{T-1}_{t=0} \Big( \log p_\phi(x_{t+1} \mid z_{t+1}) \\
    & - D_{KL}\bigl(q_\phi(z_{t+1} \mid x_{t+1}, z_t, a_t) \;\|\; p_\phi(z_{t+1} \mid z_t, a_t)\bigr) \Big) \bigg] \bigg]. \nonumber
\end{align}
For ease of notation, we have $q(z_1\mid x_1, z_0, a_0) := q(z_1\mid x_1, s)$ and $p(z_1\mid z_0, a_0) := p(z_1)$. The ELBO provides a lower bound to the log-likelihood of the observed data, which is computationally intractable to compute directly. It is composed of two terms: the expected log-likelihood of the observed data given the latent variables, and the Kullback-Leibler (KL) divergence between the variational distribution and the prior distribution of the latent variables. Minimizing the KL divergence can be interpreted as enforcing the variational distribution to be as close as possible to the prior, while maximizing the expected log-likelihood ensures that the model accurately captures the distribution of the observed data. To predict the adherence, we have $\log p_\phi(y_{t+1} \vert z_{t+1})$ as a regularisor in the loss function.

The objective is to compute the parameters 
$\phi$ that minimize the KL divergence between the variational and prior distributions of the latent variables, subject to certain constraints. These constraints are related to the expected log-likelihood of the data under the model and are represented by the inequalities with thresholds 
$\xi$. These thresholds ensure that while minimizing the losses, the model also satisfies a minimum standard for score prediction and adherence classification performances.

Latent variable models, such as Variational Autoencoders (VAEs) \citep{kingma2013,rezende14} and their variants (e.g., SLAC), often encounter challenges \citep{sonderby2016,kingma2016improved}. 
Furthermore, a higher ELBO does not always lead to enhanced predictive performance, as discussed by \citet{2018AlemiBrokeELBO, higgins2017beta}. However, the integration of scheduling strategies inspired by constrained optimization methods has been shown to significantly improve the training of latent variable models \citep{rezende2018taming,klushyn2019learning,sun2024m}. Consequently, we formulate the training of our model into an optimization problem
\begin{align}
\label{EqnConstOpt}
 \min_{\phi} 
 & 
\mathop{\mathbb{E}}_{(x_{1:T}, a_{1:T-1}) \sim D}
    \left[
    \sum^{T-1}_{t=0} 
    \left[
     D_{KL}\left(q_\phi(z_{t+1}\mid x_{t+1}, z_t, a_t)\;\|\;p_\phi(z_{t+1}\mid z_t, a_t)\right)
     \right]
     \right]
 \\
 \text{s.t.}
 &
 \quad
      \mathop{\mathbb{E}}_{(x_{1:T}, a_{1:T-1}) \sim D}
     \left[
     \mathop{\mathbb{E}}_{z_{1: T}\sim q_\phi} \left[
     \sum^{T-1}_{t=0} - \log p_\phi(x_{t+1} \mid z_{t+1})\right] \right]\leq \xi_\mathrm{score} 
 \label{eq:score}
 \\
&
     \mathop{\mathbb{E}}_{(x_{1:T}, a_{1:T-1}, y_{1:T-1}) \sim D}
     \left[
     \mathop{\mathbb{E}}_{z_{1: T}\sim q_\phi}
     \left[
     \sum^{T-2}_{t=0} 
     -\log p_\phi(y_{t+1} \mid z_{t+1})\right] 
     \right] \leq \xi_\mathrm{adherence} 
\label{eq:corss_entropy}
\end{align}
where $\xi$ is a baseline error, in Eq.~(\ref{eq:score}) we have regression with Gaussian distribution, and in Eq.~(\ref{eq:corss_entropy}) we use cross-binary entropy loss for classification.
To solve the optimization problem, we incorporate the constraints into the objective function using Lagrange multipliers $\lambda$. We apply methods from \citep{chen2022local} to adapt $\lambda$. This allows the model to balance the importance of the constraints relative to the divergence terms, which can help in avoiding common pitfalls in training such as suboptimal local minima and posterior collapse.

To avoid over-fitting, we incorporate dropout \citep{srivastava2014dropout} and Mixup \citep{zhang2017mixup}.
Subsequent research has extended the application of Mixup to latent variable models, specifically within the latent space (e.g., \citep{chen2020learning}).
However, considering our need for data augmentation across all data dimensions, not limited to latent variables, we have selected to implement the original Mixup method in our experiments.

\section{LSTM}
\label{app:lstm}

The primary objective of this study is to forecast \( y_t \) from historical data, formulated as \( y_t = f(x_{1:t}, y_{1:t-1}, s) \). To align this approach with the SLVM of SLAC for score prediction, an additional term $x_{t+1}$ is also predicted,
\begin{align}
    (x_{t+1}, y_t) = f(x_{1:t}, y_{1:t-1}, s)
\end{align}
where $f$ is a function represented by an LSTM. The loss consists of the cross entropy for adherence classification and the Normalized Mean Squared Error Loss (NMSE) for score prediction.

In our scenarios, SLVM stands out due to its inherent flexibility over traditional sequential models like LSTM. This flexibility is primarily observed in its predictive capabilities. SLVM can predict $y_t$ and use this prediction to influence the subsequent $x_{t+1}$. In contrast, LSTM only predicts a pair of $y_{t}$ and $x_{t+1}$ simultaneously, implying that we cannot use $y_{t}$ to alter $x_{t+1}$. Although it is possible to modify the LSTM model to predict a pair of $y_{t}$ and $x_{t}$, this approach encounters a similar issue for $y_t$: it cannot predict $y_t$ using the information from $x_t$.

\section{Architecture and computation}

In this study, computational experiments were performed using an NVIDIA GeForce GTX 1080 Ti GPU, with the implementation done in PyTorch, version $2.1.0$.

The SLVM model's architecture featured 32 hidden dimensions each for variables $z_1$ and $z_2$. Its encoder and decoder were symmetrically structured, each comprising five layers with 128 units. The primary activation function was LeakyReLU, set with a negative slope coefficient of 0.2. Both the encoder and decoder's mean output layers were linear, while the STD layer utilized a Softplus activation. For binary classification tasks, a Sigmoid activation was used for output.

The LSTM architecture included a hidden dimension size of 128, with two LSTM layers. The output activation function for score prediction was linear, and as in the SLVM model, a Sigmoid function was used for binary classification outputs.

Both models shared the same optimization settings. They used the RAdam \citep{liu2019variance} optimizer with a learning rate of 0.001. The batch size was set at 64, and a gradient clipping value of 0.8 was applied to ensure training stability. To prevent overfitting and enhance model generalization, a dropout rate of 0.05 was introduced. Additionally, both models incorporated Mixup as a data augmentation during training.

\end{document}